\documentclass[lettersize,journal]{IEEEtran}
\usepackage{amsmath,amsfonts}
\usepackage{algorithmic}
\usepackage{algorithm}
\usepackage{array}
\usepackage[caption=false,font=normalsize,labelfont=sf,textfont=sf]{subfig}
\usepackage{textcomp}
\usepackage{stfloats}
\usepackage{url}
\usepackage{verbatim}
\usepackage{graphicx}

\usepackage{xcolor}
\usepackage{colortbl}
\usepackage{multirow}   
\usepackage{array}      
\usepackage{booktabs}

\usepackage[colorlinks=true,linkcolor=red,citecolor=blue,urlcolor=blue,anchorcolor=blue,]{hyperref} 
\usepackage[doipre={doi:~}]{uri}
\RequirePackage{CJKnumb}
\usepackage[nocompress]{cite}  

\begin{document}

\title{Combating Pattern and Content Bias: Adversarial Feature Learning for Generalized
 AI-Generated Image Detection}

\author{Haifeng Zhang, Qinghui He, Xiuli Bi, Bo Liu~\IEEEmembership{Member,~IEEE,},\\ Chi-Man Pun~\IEEEmembership{Senior Member,~IEEE,} and Bin Xiao
\thanks{This work was supported in part by the National Science Foundation of China under Grants 62376064, 62536002, 62561160098 and 62576060, and by the National Natural Science Foundation of China Joint Key Project under Grants U24B20173, and by the Chongqing Graduate Research and Innovation Project under Grant CYB25261. (Corresponding author: Bo Liu and Bin Xiao.)}
\thanks{Haifeng Zhang and Qinghui He are with the Chongqing Key Laboratory of Image Cognition, School of Computer Science and Technology, Chongqing University of Posts and Telecommunications, Chongqing 400065, China (e-mail: D240201065@stu.cqupt.edu.cn; D250201011@stu.cqupt.edu.cn).}
\thanks{Xiuli Bi and Bo Liu are with the School of Artificial Intelligence, Chongging Key Laboratory of Image Cognition, Chongging University of Posts and Telecommunications, Chongqing 400065, China (email: bixl@cqupt.edu.cn; boliu@cqupt.edu.cn).}
\thanks{Chi-Man Pun is with the Faculty of Science and Technology of the University of Macau, Macau 999078, China (e-mail: cmpun@umac.mo).}
\thanks{Bin Xiao is with the Chongqing Key Laboratory of Image Cognition, School of Artificial Intelligence, Chongqing University of Posts and Telecommunications, Chongqing 400065, China and also with the Jinan Inspur Data Technology Co., Ltd., Jinan 250101, China (e-mail: xiaobin@cqupt.edu.cn).}}

\markboth{Journal of \LaTeX\ Class Files,~Vol.~14, No.~8, August~2021}%
{Shell \MakeLowercase{\textit{et al.}}: A Sample Article Using IEEEtran.cls for IEEE Journals}


\maketitle

\begin{abstract}
In recent years, the rapid development of generative artificial intelligence technology has significantly lowered the barrier to creating high-quality fake images, posing a serious challenge to information authenticity and credibility. Existing generated image detection methods typically enhance generalization through model architecture or network design. However, their generalization performance remains susceptible to data bias, as the training data may drive models to fit specific generative patterns and content rather than the common features shared by images from different generative models (asymmetric bias learning). To address this issue, we propose a Multi-dimensional Adversarial Feature Learning (MAFL) framework. The framework adopts a pretrained multimodal image encoder as the feature extraction backbone, constructs a real–fake feature learning network, and designs an adversarial bias-learning branch equipped with a multi-dimensional adversarial loss, forming an adversarial training mechanism between authenticity-discriminative feature learning and bias feature learning. By suppressing generation-pattern and content biases, MAFL guides the model to focus on the generative features shared across different generative models, thereby effectively capturing the fundamental differences between real and generated images, enhancing cross-model generalization, and substantially reducing the reliance on large-scale training data. Through extensive experimental validation, our method outperforms existing state-of-the-art approaches by 10.89\% in accuracy and 8.57\% in Average Precision (AP). Notably, even when trained with only 320 images, it can still achieve over 80\% detection accuracy on public datasets.
\end{abstract}


\section{Introduction}
\label{sec:intro}

\IEEEPARstart{T}{he} rapid development of generative artificial intelligence has significantly lowered the barrier to creating high-quality generated images~\cite{keita2025fidavl},\cite{fu2025faces}. With the continuous advancement of deep generative models, AI-generated image technology has garnered widespread attention~\cite{effort2024effort},\cite{zheng2024breaking}. Even individuals without professional technical knowledge can easily generate highly realistic images. While this presents tremendous opportunities for fields such as visual arts, it also poses significant risks, including the spread of misinformation~\cite{epstein2023online},\cite{prashnani2024generalizable}. Therefore, there is an urgent need to develop AI-generated image detection frameworks capable of effectively distinguishing between real and generated images.

Earlier detection methods~\cite{20chai2020makes,18wang2020cnn,6marra2019gans,yu2019attributing} 
primarily focused on identifying specific artifacts left by generative models in the spatial or frequency domains, such as grid artifacts introduced during network upsampling. These methods achieve strong performance when detecting images generated by traditional GANs or early diffusion models, but these methods exhibit significant performance degradation when confronted with images produced by recently popular generative models that belong to different model families from those used during training. Meanwhile, diffusion-reconstruction-based detection methods~\cite{26wang2023dire,32luo2024lare} demonstrate high accuracy in detecting diffusion-generated images, yet similarly suffer from notable performance drops when applied to GAN-generated images. As deep generative models continue to rapidly evolve, 
the core challenge faced by current generative image detectors has increasingly shifted toward generalization ability~\cite{ucf2023ucf,3bias}, 
especially when dealing with generative models unseen during training.

Inspired by the substantial improvements in generalization performance brought by large-scale pretrained multimodal models in image classification tasks, recent studies~\cite{2ojha2023towards},\cite{CLIPping2024clipping},\cite{effort2024effort},\cite{peng2025face} have begun leveraging such pretrained multimodal models for generative image detection, achieving better generalization than traditional approaches. However, these methods remain fundamentally data-driven. Given the high diversity in generative patterns and content within generative image detection, the training data itself plays a crucial role in determining generalization ability. Nonetheless, existing approaches tend to focus on optimizing model architectures or exploring different image representation domains in hopes of capturing more general and fine-grained discriminative features, while often overlooking the potential generalization limitations introduced by the training data.

\begin{figure*}[t]
	\centering
	\includegraphics[width=0.88\linewidth]{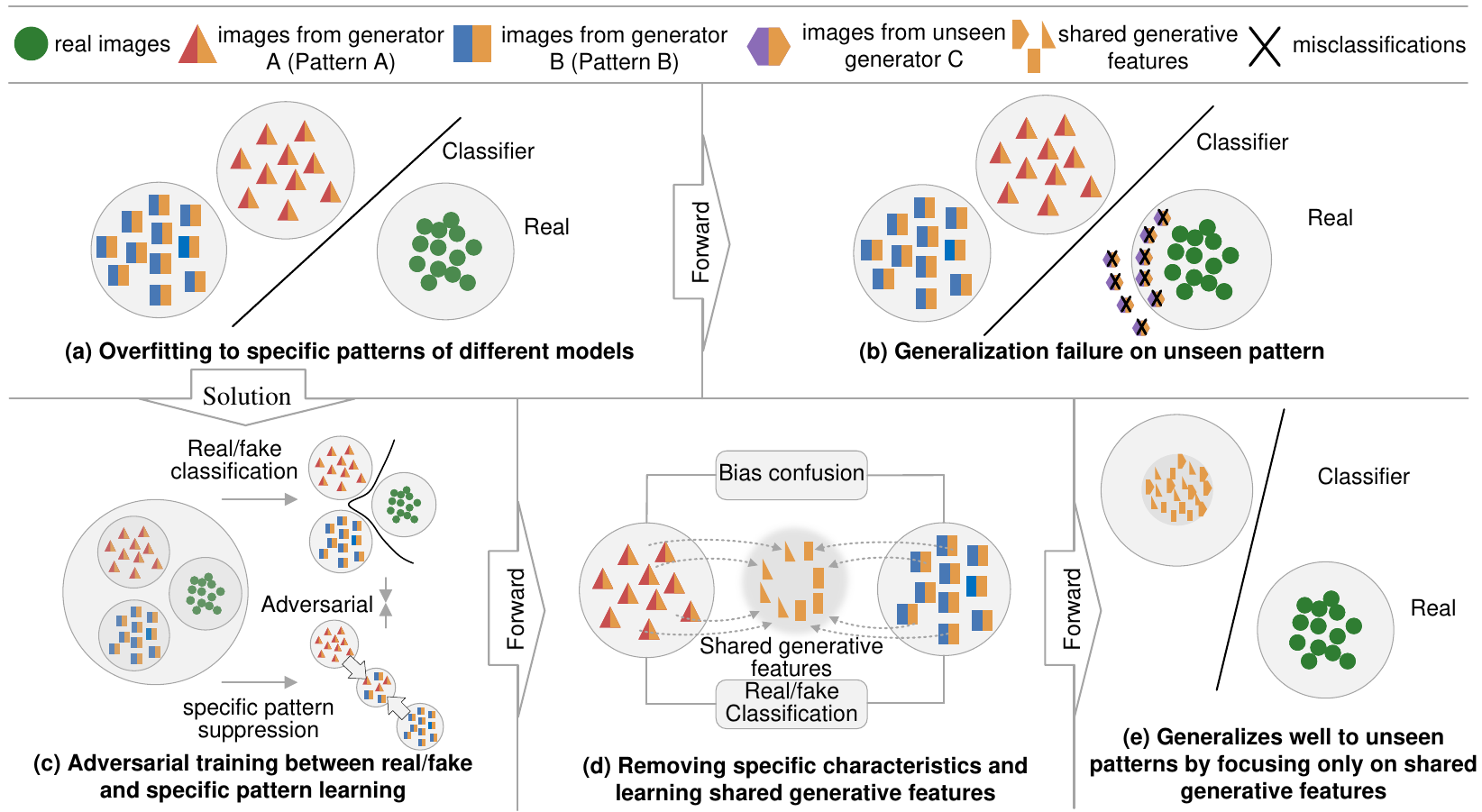}
	\caption{(Color best viewed.) This illustration shows the problems of traditional methods under the asymmetric bias learning phenomenon and our proposed solution. (a) Traditional detectors (e.g., CLIP + classifier) tend to over-rely on the specific generative patterns of particular generative models when performing real/fake classification. (b) Such reliance on specific generative patterns makes the model prone to generalization failure when encountering unseen generative patterns. (c)–(e) MAFL introduces an adversarial mechanism between the real/fake classification task and the specific-pattern classification task, guiding the model to extract common generative features that characterize different generative models. These features are free from model-specific characteristics, thereby enabling stronger generalization when facing unseen generative models.}
	\label{FIG:1}
\end{figure*}

Through an in-depth analysis of the above approaches, we identify a common generalization bottleneck induced by the training data. From the perspectives of generative patterns and image content, this limitation can be primarily attributed to two aspects:
(1) Generative pattern bias: Due to differences in training data, internal network architectures, and generative mechanisms, different generative models typically exhibit distinctive generative artifacts and inherent stylistic tendencies, such as artistic, cinematic, or photorealistic styles. During detector training, the limited diversity of generative models in the training set often causes the detector to over-rely on the specific generative patterns associated with the seen models as primary decision cues. As a result, the detector may fail to correctly classify images generated by models with different generative patterns.
(2) Content bias: Instead of learning the fundamental generative characteristics that distinguish real images from generated ones, detectors tend to exploit simpler but less generalizable features—such as semantic content—as shortcuts for classification, which leads to erroneous predictions. For example, when a real image contains content that frequently appears in generated images within the training set, the detector may incorrectly classify it as a generated image. We collectively refer to these phenomena as asymmetric bias learning. As shown in Fig. \ref{FIG:1}(a)-(b), this problem severely undermines the generalization ability of existing generated image detection models.

To correct this undesirable behavior, we enforce the model to learn the common features shared by images from different generative models—features that transcend specific patterns and content and are crucial for distinguishing between real and generated images. As a concrete implementation of this concept, we introduce a novel approach called the MAFL. This framework leverages an adversarial training mechanism to prevent the model from overfitting to generative patterns and content, thereby avoiding asymmetric bias learning, as shown in Fig. \ref{FIG:1}(c)-(e). Specifically, we first extract image features using a large-scale pretrained multimodal model. Then, by establishing an adversarial relationship between the real/fake classifier and the bias learner, the bias network continuously improves its ability to identify patterns and content biases, while the real/fake classifier attempts to guide the feature extractor to learn features that can bypass the bias network but remain effective for real/fake discrimination. We constrain this process from three complementary perspectives: probability distribution, feature space, and downstream tasks. This framework effectively guides the model to focus on the common features shared across images generated by different models, thereby enhancing its generalization ability to unseen generative models.

In summary, our contributions are as follows:

\begin{enumerate}
  \item We introduce asymmetric bias learning, where training data drive detectors to overfit to specific generative patterns and content, limiting generalization to unseen models. By framing bias as an adversary, we force the model to rely only on generative features shared across different generative models.
  \item To this end, we propose the MAFL framework, which introduces an adversarial game between a bias-learning network and a real–fake discriminator, together with a multi-dimensional adversarial loss. This design encourages the model to learn generative features shared across different generative models and mitigates overfitting to specific patterns or content.
  \item Some existing methods mainly address content bias and rely on content alignment during dataset construction, while largely overlooking generative pattern bias. In contrast, our method directly suppresses both content bias and generative pattern bias at the feature level within the network, without requiring any manual dataset modification or reconstruction.
\end{enumerate}

The remainder of this paper is organized as follows. Section \ref{sec:Related work} reviews related work, with a particular focus on recent advances addressing generalization and data bias issues in generated image detection. Section \ref{sec:approach} presents the motivation of this study and provides a detailed description of the proposed MAFL framework, including its adversarial mechanism and multi-dimensional loss design. Section \ref{sec:Experiments} reports performance comparisons to verify the effectiveness of the proposed method. Section \ref{sec:Conclusion} concludes the paper.

\section{Related Work}
\label{sec:Related work}

In this section, we first review the existing methods for generated image detection, and then discuss approaches that address issues of generalization and data bias.

\subsection{AI-generated Image Detection}
With the rapid development of AI-based image generation technologies, a wide range of detection methods have been proposed. CNNSpot~\cite{18wang2020cnn} found that a CNN combined with two operations (Gaussian blur and JPEG compression), when trained on ProGAN, can successfully detect images generated by other GANs. FreDect~\cite{23frank2020leveraging} showed that GAN-generated images exhibit pronounced artifacts in the frequency domain, which can serve as effective discriminative cues. Liu et al.~\cite{24liu2022detecting} observed that noise patterns in real images display consistent characteristics in the frequency domain, whereas generated images differ from them. This discrepancy is thus exploited to train classifiers and improve detection performance. Bi et al.~\cite{25bi2023detecting} approached the problem from a frequency-domain perspective by mapping real images into a specific subspace, thereby enhancing the model’s generalization ability. Lgrad~\cite{27tan2023learning} transformed images into the gradient domain and trained classifiers using artifacts revealed in the gradients. DIRE~\cite{26wang2023dire} introduced diffusion reconstruction error, which performs classification and detection based on reconstruction error between real and generated images.

More recent studies have approached performance improvement from new visual perspectives. NPR~\cite{npr2024rethinking} analyzed local pixel distribution patterns during upsampling at the pixel level and used neighboring pixel relationships as generative artifacts for detection. CLIPping~\cite{CLIPping2024clipping} adopted a lightweight prompt-tuning–based adaptation strategy to enhance CLIP’s capability for detecting generated images. Effort~\cite{effort2024effort} expanded the feature space via singular value decomposition to mitigate overfitting to spurious features.

Overall, existing methods achieve high accuracy when detecting images generated by models seen during training, but their performance often degrades significantly when confronting unseen generative models, especially those from different model families. Given the continuous evolution of image generation techniques and the constant emergence of new generative models, the generalization capability of detectors has become a central focus of current research.

\subsection{Generalization Performance}
To improve the generalization of generative image detection, researchers typically refine model architectures or explore alternative image representations to extract more fine-grained or generalizable features. CNNSpot~\cite{18wang2020cnn} explored the generalization performance of convolutional neural network-based detectors in generated image detection tasks. Zhao et al.~\cite{21zhao2021multi} modeled the detection task as a fine-grained classification problem to extract more detailed image features and enhance generalization. Fusing~\cite{19ju2022fusing} designed a dual-branch framework that combines global and local image features to enhance detection accuracy. Meanwhile, the strong transferability of large-scale pretrained models has also attracted considerable attention~\cite{fan2024scaling},\cite{cozzolino2024zero},\cite{de2024exploring}. Univfd~\cite{2ojha2023towards} utilized CLIP’s image encoder to extract features for detecting various generative models, while \cite{1cozzolino2024raising} pointed out that detectors based on large models may rely on high-level semantic features, which further strengthens generalization.

Despite these advances, most existing methods pay little attention to the influence of data itself on detection performance. Given the data-driven nature of deep models and the complex, diverse content and patterns of real and generated images, current methods often fail to generalize to unseen content or patterns.

\subsection{Asymmetric Bias Learning}
 Recent studies have conducted preliminary explorations into content bias. \cite{1cozzolino2024raising} constructed a training set with one-to-one content correspondence between real and generated images, while \cite{guillaro2024bias} used text prompts to ensure close content alignment between generated and real images, reducing bias caused by semantic gaps. However, these methods face two major limitations: first, they typically focus only on content bias while overlooking the impact of pattern bias. Second, these methods primarily address the problem at the dataset level rather than modeling it at the feature level, which limits their scalability and generalization. Removing only one type of bias may still lead the model to learn specific content or pattern, and methods relying on dataset-specific constraints are difficult to extend to broader scenarios, thereby restricting their generalization ability.

Unlike these methods, we innovatively employ adversarial learning to guide the model to learn the common features of images generated by different generative models, while reducing its dependence on specific content and pattern, thereby making it more sensitive to the fundamental differences between real and generated images. This approach enhances the model's generalization ability when dealing with diverse content and various generative models.

\section{Approach}
\label{sec:approach}

In this section, we first investigate the intrinsic challenges of generated image detection and present empirical evidence supporting the existence of asymmetric bias learning. Based on these insights, we introduce the proposed Multi-dimensional Adversarial Feature Learning (MAFL) framework, including its design principles, network architecture, adversarial mechanism, and objective functions.

\subsection{Observations and Motivations}
\label{Observations and Motivations}

In recent years, as the visual quality of generative models has continued to improve, the differences between real and generated images have become increasingly subtle, causing existing detectors to rely on spurious correlations during training. Most existing methods formulate generated image detection as a binary classification task~\cite{2ojha2023towards,18wang2020cnn} and follow the prevailing paradigm of employing large-scale multimodal models (e.g., CLIP) to extract image features before training a binary classifier. Given the central role of these features in determining detection performance, we conduct a systematic analysis of their feature representations and identify two key types of bias that substantially affect the generalization ability of current detectors: content bias and generative pattern bias.

\begin{figure}[t]
	\centering
		\includegraphics[width=0.85\linewidth]{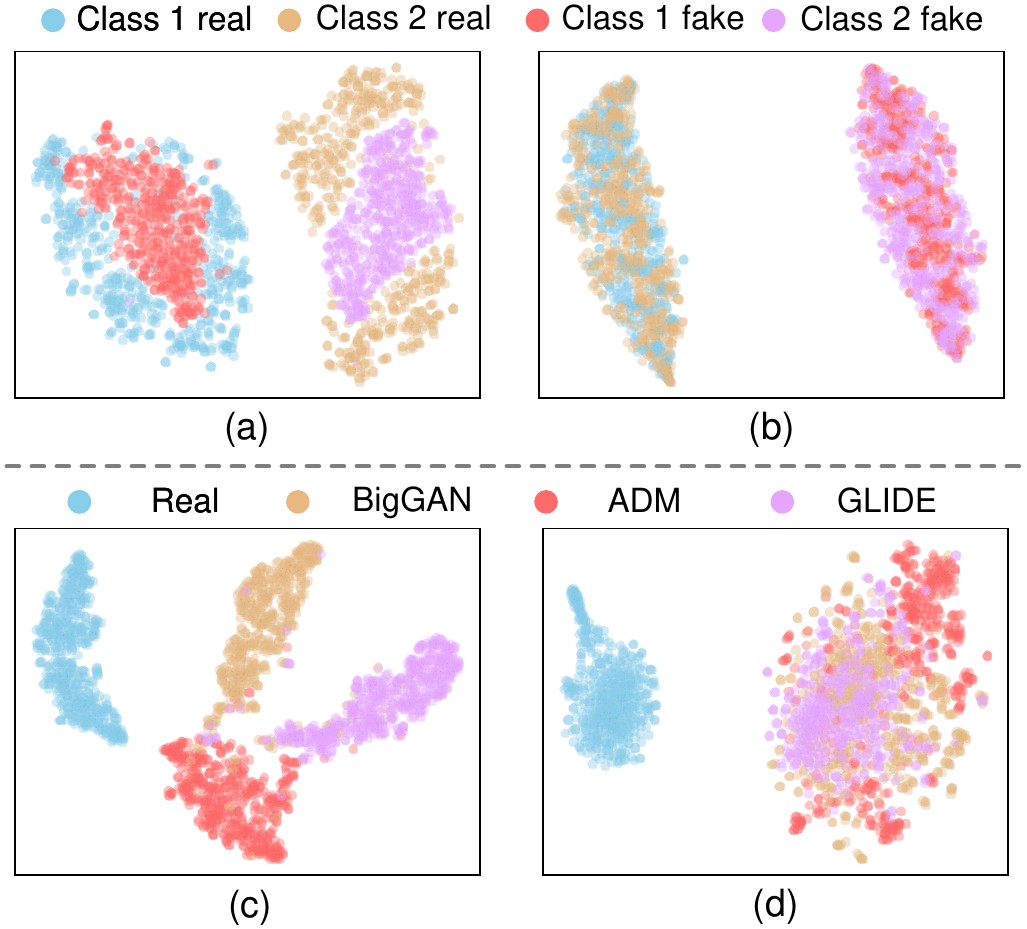}
	\caption{t-SNE visualization of features. (a) Visualization of general features extracted by CLIP, labeled by category and authenticity (real/fake). (b) Visualization of features extracted by MAFL, labeled by category and authenticity (real/fake). (c) Visualization of CLIP features mapped to MLP, trained using real and generated images(fake). (d) Visualization of mapped features after Multi-dimensional Adversarial Feature Learning.}
	\label{FIG:2}
\end{figure}

\textit{1)Content Bias:} We first define content bias as the tendency of a detector to rely on semantic content, scene composition, or other content-related spurious cues that correlate with authenticity, rather than learning the intrinsic differences between real and generated images. This issue is particularly critical because generated image detection must operate in diverse real-world scenes whose content distributions differ drastically from limited training data. To verify this bias, we visualize CLIP features using t-SNE. As shown in Fig.~\ref{FIG:2}(a), the feature space splits into two distinct clusters. However, these clusters align with semantic categories rather than real/fake labels. This demonstrates that CLIP embeddings are largely structured by semantic information, leading detectors to exploit semantic differences for classification and introducing undesired content-induced interference into authenticity prediction.

\begin{figure}[t]
	\centering
		\includegraphics[width=1\linewidth]{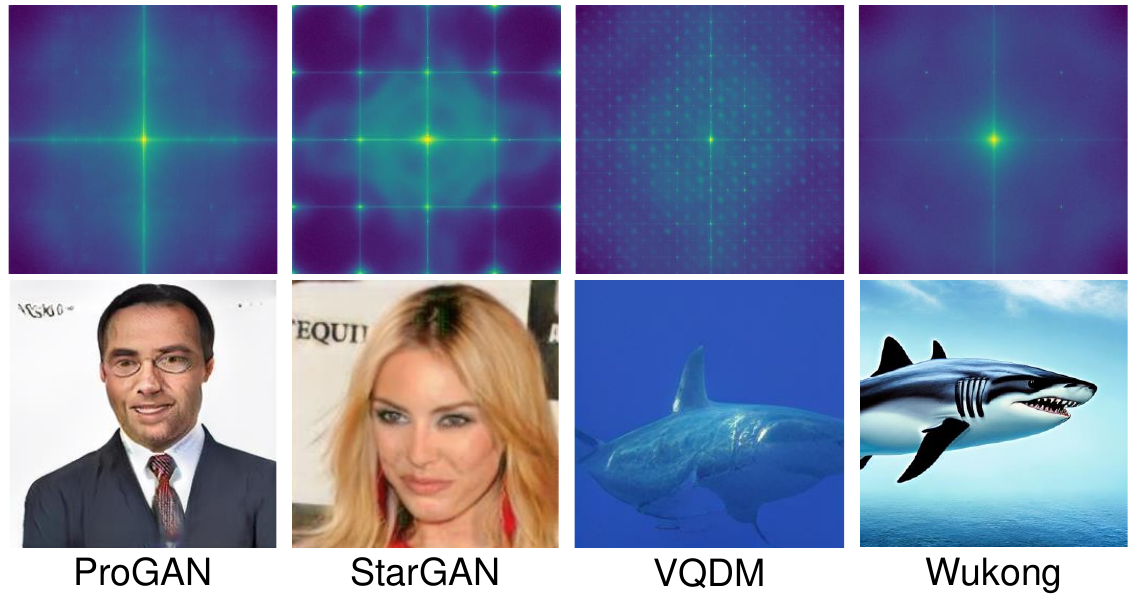}
	\caption{The average frequency spectra and sample examples of images generated by different models are shown. From the frequency domain perspective, we can clearly observe that each model exhibits distinctive generation artifacts. Additionally, at the image level, they display noticeable style differences.}
	\label{FIG:3}
\end{figure}

\textit{2)Generative Pattern Bias:} Generative models encompass a wide range of architectures, each differing in generation mechanisms, internal structures, and training data. As a result, different models tend to exhibit unique frequency-domain patterns, characteristic artifacts, or default visual styles. We define generative pattern bias as the tendency of a detector to rely on such model-specific attributes—rather than the fundamental generative features shared across models—to perform real/fake discrimination. To validate the existence of this bias, we fine-tune CLIP for real–fake classification~\cite{2ojha2023towards} and visualize the resulting feature representations. As shown in Fig.~\ref{FIG:2}(c), generated images from different models form clearly separated clusters, indicating strong model-specific generative patterns. Combined with the frequency-domain evidence in Fig.~\ref{FIG:3}, which reveals markedly distinct spectral structures across different generative families (e.g., diffusion models vs. GANs), these observations show that detectors relying on such patterns are prone to failure when encountering unseen generative models.

\textit{3)Asymmetric Bias Learning:} Based on the above definitions and empirical observations, we summarize these behaviors as asymmetric bias learning. (1) Content-side asymmetry: the detector relies on incidental semantic differences between real and generated images in the training data instead of true authenticity cues. (2) Pattern-side asymmetry: the detector overfits to the generative patterns of seen models, resulting in poor discrimination on unseen models. These two forms of bias jointly degrade generalization performance. As illustrated in Fig.~\ref{FIG:2}(b)(d), when these biases are effectively suppressed, the detector can learn more generalizable features tied to the intrinsic differences between real and generated images, yielding clearer and more reliable decision boundaries. This analysis indicates that a detector with strong generalization capability must suppress both types of bias and instead learn the generative features that are shared across diverse models. This forms the theoretical foundation of our proposed MAFL framework.

\label{sec:Approach}

\begin{figure*}[t]
	\centering
	\includegraphics[width=1\linewidth]{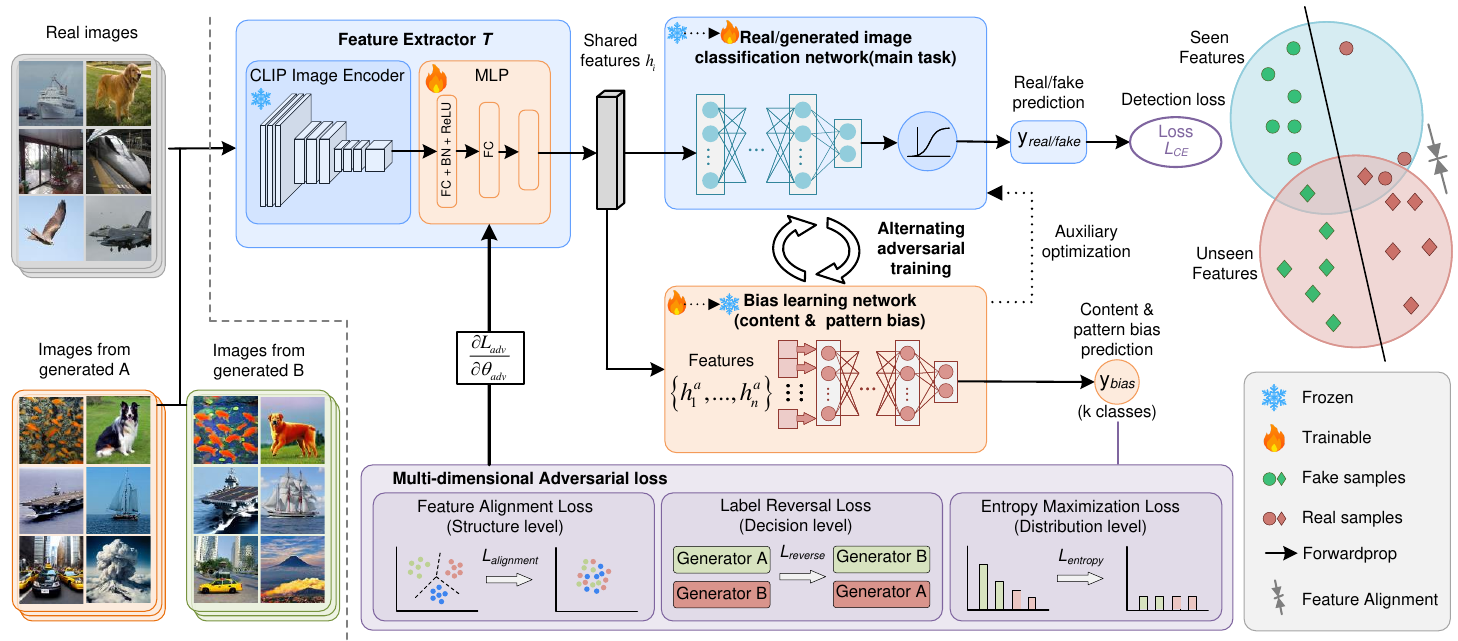}
	\caption{\textbf{MAFL Overall Structure.} Our framework consists of a feature extractor, a real/generated image classification network, and a bias learning network. Through alternating adversarial training, the bias learning network learns and distinguishes the generative pattern and content differences across various generative models. Meanwhile, the real/generated image classification network guides the feature extractor to learn shared generative features that can “deceive’’ the bias learning network and remain consistent across different generative models.}
	\label{FIG:4}
\end{figure*}

\subsection{Overall Framework}

To address the asymmetric bias learning problem, we propose a Multi-dimensional Adversarial Feature Learning (MAFL) framework. As shown in Fig.~\ref{FIG:4}, our method consists of three core components: (1) Feature Extractor: A CLIP image encoder followed by an MLP, producing shared image representations. (2) Real/Generated Image Classification Network: Learns discriminative features for authenticity classification. (3) Bias Learning Network: Identifies generative-pattern biases across different generative models, as well as content biases within the real and generated image classes.

The key idea is to establish an alternating adversarial relationship between the real/generated image classifier and the bias learner. The bias learning network actively mines both pattern and content biases, while the real/fake classifier and feature extractor collaborate to suppress such biases and retain only authenticity-relevant signals. Through this adversarial pressure, the model progressively eliminates spurious dependencies and focuses on shared generative features.

\subsection{Network Architecture}

\textit{1)Feature Extractor:} Our feature extractor $T$ consists of a CLIP image encoder and an MLP. For an input image pair $({x_0},{x_1})$, where ${x_0}$ represents a real image and ${x_1}$ represents a generated image, the feature extraction process can be expressed as:
\begin{equation}
{h_i} = {\rm{ML}}{{\rm{P}}_g}({\rm{CLIP}}({x_i})){\rm{, }}i \in \left\{ {0,1} \right\},
  \label{eq:1}
\end{equation}
where ${h_i}$ represents the extracted class-specific features.

\textit{2)Real/Generated Image Classification Network:} Based on $h_i$, the real/fake classifier predicts authenticity:
\begin{equation}
{y_{{\rm{real/fake}}}} = {\rm{ML}}{{\rm{P}}_{{\rm{real/fake}}}}({h_i}),i \in \left\{ {0,1} \right\},
\label{eq:2}
\end{equation}

\textit{3)Bias Learning Network:} The bias network focuses on uncovering generative pattern biases in generated images, as well as content biases within both real and generated images, taking the corresponding feature representations $\left\{ {h_1^a,h_2^a,...,h_n^a} \right\}$ as input:
\begin{equation}
{y_{{\rm{bias}}}} = {\rm{ML}}{{\rm{P}}_{{\rm{bias}}}}(h_i^a).
\label{eq:3}
\end{equation}

This auxiliary task provides adversarial pressure to the feature extractor, as the extractor must learn representations that hinder bias prediction while preserving real/fake separability.

\subsection{Adversarial Feature Learning}

Our proposed adversarial training mechanism establishes a “game” relationship within the model: the bias learning network is responsible for identifying model-specific characteristics of different generative models as well as content-related features of images, while the real/generated image classification network guides the feature extractor to produce representations that can effectively deceive the bias learning network. This mechanism consists of two alternating phases:

\textit{1)Bias learning phase:} After feature extraction, images are fed into the bias learning network, which is trained to distinguish model-specific generative patterns as well as content-related features. During this phase, only the parameters of the bias learning network are updated, while the feature extractor and the real/generated image classification network remain fixed.

\textit{2)Adversarial learning phase:} The same image features are simultaneously fed into the real/generated image classification network and the pretrained bias learning network with frozen parameters. Guided by the adversarial loss function, the feature extractor and the real/generated classification network are jointly optimized to learn feature representations that are insensitive to generative patterns and semantic content, while remaining effective for real/generated discrimination.

During training, unlike conventional domain-adversarial learning where the two branches play symmetric adversarial roles, the bias learning phase in our framework is designed as a purely auxiliary task for mining and amplifying bias features. Its objective is to continuously impose adversarial pressure on the feature extractor, encouraging it to gradually eliminate its dependence on generator-specific patterns and semantic biases while preserving real/generated discrimination capability. This auxiliary task alternates with the main task, forming a dynamic equilibrium: the bias learning network continuously improves its ability to recognize generative patterns and semantic biases, while the feature extractor persistently learns to produce features that deceive the bias network yet retain strong real/generated discrimination ability.

\subsection{Objective Function}
To effectively suppress the learning of content and generative pattern biases, we propose a multi-dimensional adversarial loss that imposes complementary constraints on feature learning. These constraints operate from three distinct perspectives, guiding the model to focus on intrinsic differences between real and generated images:

\begin{equation}
{{\cal L}{adv}} = {{\cal L}{entropy}} + \alpha \cdot {{\cal L}{alignment}} + \beta \cdot {{\cal L}{reverse}}
\label{eq:adv}
\end{equation}

where $\alpha$ and $\beta$ are weighting coefficients that balance the respective loss terms.

\noindent
\textbf{Entropy Maximization Loss.}
This loss regulates feature learning from the probability distribution perspective. By enforcing the KL divergence~\cite{kullback1951information} between the pattern prediction distribution and a uniform prior, the feature extractor is encouraged to erase generator-dependent statistical differences and maximize uncertainty in pattern prediction:

\begin{equation}
{{\cal L}_{entropy}} = D_{KL}({\rm{softmax(z)}} \parallel \frac{1}{K}\mathbf{1})
  \label{eq:entropy}
\end{equation}
where $z$ denotes the logits of the bias learning network, $K$ is the number of generator types, and $\frac{1}{K}\mathbf{1}$ represents the uniform distribution.

\noindent
\textbf{Feature Alignment Loss.}
From the feature-structure perspective, this loss enforces geometric consistency among forged features by maximizing their pairwise similarity. It encourages forged images to cluster on a unified forgery manifold:

\begin{equation}
\mathcal{L}_{\text{alignment}} = \frac{1}{N(N-1)}\sum_{i \neq j}(1 - S_{ij})
\label{eq:alignment}
\end{equation}
where $S_{ij} = f_i \cdot f_j^T$ denotes the cosine similarity between normalized features $f_i$ and $f_j$, and $N$ is the batch size. By minimizing the average dissimilarity among generated samples, this loss aligns their representations while preserving necessary diversity to avoid degenerate feature collapse

\noindent
\textbf{Label Reversal Loss.}
From the decision-level perspective, this loss assigns an adversarial target opposite to the true generator label, preventing the model from leveraging pattern-specific cues. It explicitly pushes the feature extractor away from pattern-sensitive directions:
\begin{equation}
{{\cal L}_{reverse}} =  - \sum\limits_{i = 1}^N \sum\limits_{k = 1}^K {(1 - {y_{i,k}})} \log {p_{i,k}}
  \label{eq:reverse}
\end{equation}
where $y_{i,k}$ is the one-hot encoded generator label and $p_{i,k}$ is the softmax probability predicted by the bias learning network.

Through the integration of these three complementary constraints, the proposed adversarial objective effectively suppresses generator-specific biases from distribution, structure, and decision perspectives, enabling the model to learn generator-agnostic representations and significantly improving generalization on unseen generative models.

\section{Experiments}
\label{sec:Experiments}

This section first presents the experimental settings, including the datasets, evaluation metrics, comparison methods, and implementation details. Subsequently, we conduct a comprehensive evaluation and analysis of the proposed method through comparison experiments, ablation studies, robustness analysis, and experiments under limited-sample scenarios. Finally, we summarize and discuss the current limitations of the method.

\subsection{Experimental Setup}

\textit{1)Datasets:} To comprehensively evaluate the proposed method across diverse scenes and different generative paradigms—including diffusion models, GANs, and autoregressive models—we conduct systematic experiments on 25 generative models sourced from three representative public datasets: Holmes\cite{holms2025aigi}, ForenSynths\cite{18wang2020cnn}, and GenImage\cite{genimage2024genimage}. We design the following three main evaluation protocols:

(1) Protocol-1.We first evaluate the performance of the proposed method on the latest, previously unseen models. We train and test on the Holmes dataset, where the training set consists of 65,000 images in total, including real images and those generated by multiple diffusion models. The test set covers state-of-the-art diffusion models FLUX\cite{flux2024}, PixArt-XL\cite{pixart2024pixart}, SD3.5\cite{sd352024scaling}, as well as recent autoregressive generative models LlamaGen\cite{Llamagen2024autoregressive}, Infinity\cite{Infinity2025infinity}, Janus-Pro\cite{januspro2025janus}, VAR\cite{VAR2024visual}, Janus\cite{janus2025janus}, and Show-o\cite{show2024show}. Each test subset contains 5,000 real images from the COCO dataset\cite{COCOlin2014microsoft} and 5,000 fake images generated by the corresponding model.

(2) Protocol-2.To assess the method’s detection capability across model families (from GANs to diffusion models), we train on ProGAN and StyleGAN, two commonly used generators in the ForenSynths dataset. The training set consists of approximately 20,000 real and generated images, covering 23 scene categories. The trained models are then evaluated on both the ForenSynths and GenImage test sets. The ForenSynths test set, which is primarily GAN-based, includes subsets generated by ProGAN, CycleGAN, StyleGAN, GauGAN, Deepfake, and SAN. The corresponding real images are drawn from six datasets: LSUN\cite{lsunyu2015lsun}, ImageNet\cite{imagenetrussakovsky2015imagenet}, CelebA\cite{celebaliu2015deep}, CelebA-HQ\cite{progankarras2017progressive}, COCO\cite{COCOlin2014microsoft}, and FaceForensics++\cite{facerossler2019faceforensics++}, with a total of about 46,000 real and generated images.

(3) Protocol-3. To evaluate the proposed method’s detection capability on other unseen models after being trained on diffusion models, we use images generated by Stable Diffusion v1.4 and VQDM—the two most common training subsets in the GenImage dataset—along with their corresponding real images for training. The trained models are then evaluated on the GenImage and ForenSynths test sets, which together cover 15 generative models, including diffusion models, GANs, and Deepfake methods.
This protocol, together with Protocol-2, forms a cross-evaluation setting (GAN → diffusion and diffusion → GAN), enabling us to investigate the proposed method’s cross-family generalization ability.

\begin{figure*}[htbp]
	\centering
		\includegraphics[width=0.95\linewidth]{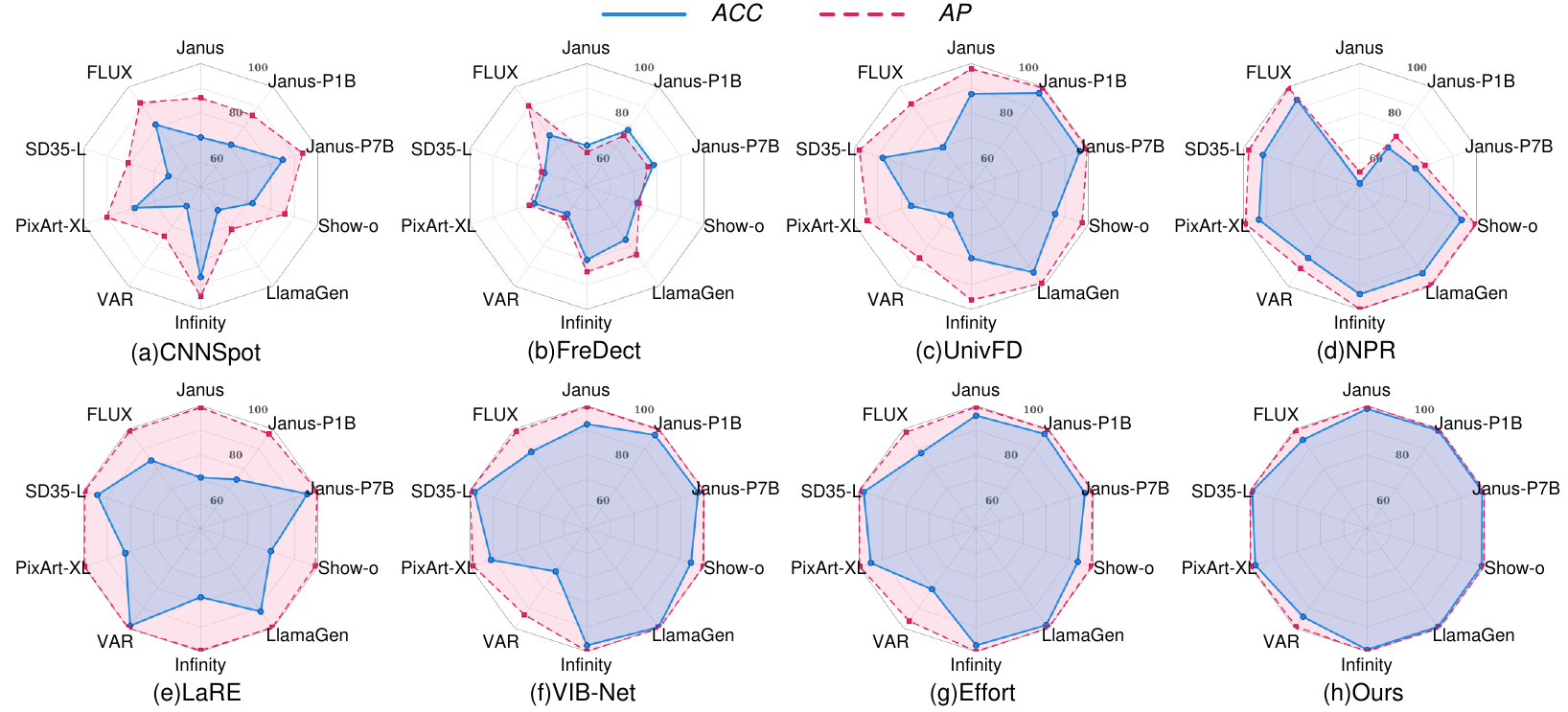}
	\caption{Performance evaluation on the latest unseen generative models.All methods are evaluated under Protocol-1 for a fair comparison. Our method achieves superior detection performance on the latest unseen generative models compared with existing approaches.}
	\label{fig5_radar_charts}
\end{figure*}

\textit{2)Evaluation Metrics:} We evaluate our method using four commonly adopted metrics: Accuracy (\textit{ACC}), Average Precision (\textit{AP}), \textit{F1} score, and Area Under the Curve (\textit{AUC}).

Accuracy measures the proportion of correctly classified samples:
\begin{equation}
\mathrm{\textit{ACC}} = \frac{TP + TN}{TP + TN + FP + FN}.
\end{equation}

The \textit{F1} score balances precision and recall and is defined as
\begin{equation}
\mathrm{\textit{F1}} = \frac{2 \cdot \mathrm{Precision} \cdot \mathrm{Recall}}
{\mathrm{Precision} + \mathrm{Recall}}.
\end{equation}

where Precision and Recall are given by
\begin{align}
\mathrm{Precision} &= \frac{TP}{TP + FP}, \\
\mathrm{Recall} &= \frac{TP}{TP + FN}.
\end{align}

Average Precision (\textit{AP}) summarizes the area under the Precision--Recall curve across different thresholds, while \textit{AUC} measures the area under the ROC curve, reflecting the model’s ability to distinguish real and generated images.

\textit{3)Comparison Methods:} We compared the method proposed in this research with various state-of-the-art detection techniques, including FreDect (ICML2020)~\cite{23frank2020leveraging}, CNNDetection (CVPR2020)~\cite{18wang2020cnn}, Fusing (ICIP2022)~\cite{19ju2022fusing}, LGrad (CVPR2023)~\cite{27tan2023learning}, Univfd (CVPR2023)~\cite{2ojha2023towards}, DIRE (ICCV2023)~\cite{26wang2023dire}, CLIPping (ICMR2024)~\cite{CLIPping2024clipping}, NPR (CVPR2024)~\cite{npr2024rethinking}, LaRE (CVPR2024)~\cite{LaRE2024lare}, VIB-Net (CVPR2025)~\cite{VIB2025towards}, and Effort (ICML2025)~\cite{effort2024effort}. To ensure fair comparison, all methods are evaluated under the same three protocols described above.

\textit{4)Implementation Details:}
Our model is implemented in PyTorch 2.7.1 and trained on an NVIDIA RTX A6000 GPU. 
Both the feature extractor and the bias classifier are optimized using AdamW with a learning rate of 0.0002 and a weight decay of $1\times 10^{-2}$. 
We adopt a ReduceLROnPlateau scheduler with a decay factor of 0.5 and a patience of 5 epochs. The batch size is set to 256. Training lasts for 100 epochs with early stopping (patience = 10). For adversarial training, we first pre-train the bias classifier for 5 epochs. During adversarial optimization, the bias classifier is updated three times per batch, whereas the feature extractor is updated once. The adversarial loss weight $\lambda$ is dynamically scheduled as:
\begin{equation}
\lambda = \min \left( 0.5,\; 0.1 \times \frac{\mathrm{epoch}}{10} \right).
\end{equation}
Empirically, in computing the adversarial loss, we set the feature alignment coefficient $\alpha$ to 0.5 and the label reversal coefficient $\beta$ to 0.3.

\subsection{Performance comparisons}

\textit{1)Performance Evaluation on the Latest Unseen Generative Models:} Following Protocol-1, we conduct a systematic comparison between our proposed method and several state-of-the-art deepfake detection approaches. Fig.~\ref{fig5_radar_charts} reports the performance of eight detectors on ten generative models; these ten models differ from the training distribution and represent recent diffusion and autoregressive generators.

CNNSpot employs a conventional convolutional backbone (e.g., ResNet) for feature extraction. Although such networks perform strongly on semantic classification tasks, their ability to capture subtle forgery traces in generated images is limited. FreDect detects GAN-specific artifacts in the frequency domain, which hinders its generalization to modern autoregressive and diffusion-based models. UnivFD directly uses CLIP features for classification, making it vulnerable to interference from generation-pattern and semantic content variations. NPR is designed to learn local pixel relationships: it performs well on many generators but struggles on autoregressive models where local pixel correlations are less pronounced.

VIB-Net and Effort both aim to refine the CLIP feature space: the former mainly suppresses irrelevant semantic features but overlooks generative-pattern biases, while the latter expands CLIP representations via orthogonal subspaces to mitigate overfitting—yet it still risks learning generator-specific patterns. By contrast, our method employs adversarial training to prevent overreliance on semantic content and generative-pattern biases, encouraging the model to learn more universal forgery fingerprints shared across different generators. Fig.~\ref{fig5_radar_charts}(h) presents our method’s detection results on the latest diffusion and autoregressive models, demonstrating that our approach achieves the best cross-model-series generalization.

\begin{table*}[t]
    \centering
    \renewcommand{\arraystretch}{1}
    \setlength{\tabcolsep}{2.5pt}
    \caption{Performance metrics (Average Precision) for various approaches trained using GAN-generated imagery. Other generative frameworks beyond the GANs can be interpreted as the domain for generalization testing.}    
    \begin{tabular}{ccccccccccccccccc}
    \toprule
    \multicolumn{1}{c}{\multirow{2}[4]{*}{Method}} & \multicolumn{6}{c}{Generative Adversarial Networks} & \multicolumn{7}{c}{Diffusion Models} & \multicolumn{2}{c}{Others} & \textit{AP} \\
    \cmidrule(r){2-7} \cmidrule(r){8-14} \cmidrule(r){15-16} \cmidrule(r){17-17}    
    & \multicolumn{1}{p{2.5em}}{Pro-GAN} & \multicolumn{1}{p{2.5em}}{Style-GAN} & \multicolumn{1}{p{2.5em}}{Cycle-GAN} & \multicolumn{1}{p{2.5em}}{Star-GAN} & \multicolumn{1}{p{2.5em}}{Big-GAN} & \multicolumn{1}{p{2.5em}}{Gau-GAN} & SDV1.4 & VQDM & SDV1.5 & GLIDE & ADM & Midj & Wukong & \multicolumn{1}{p{2.5em}}{SAN} & \multicolumn{1}{p{2.5em}}{Deep-fake} & Avg \\
    \midrule
    FreDect\cite{23frank2020leveraging}{\tiny{} (ICML2020)} & \textbf{100.00} & \textbf{100.00} & 64.29 & 99.18 & 77.16 & 74.17 & 51.14 & 72.08 & 51.50 & 40.70 & 41.58 & 46.24 & 50.36 & 45.75 & 52.53 & 64.45 \\
    CNNSpot\cite{18wang2020cnn}{\tiny{} (CVPR2020)} & \textbf{100.00} & \textbf{100.00} & 88.81 & 98.13 & 77.57 & 86.39 & 55.06 & 65.02 & 55.52 & 56.40 & 68.37 & 46.82 & 51.82 & 59.33 & 65.65 & 71.66 \\
    Lgrad\cite{27tan2023learning}{\tiny{} (CVPR2023)} & \underline{99.97} & \textbf{100.00} & 58.21 & 99.93 & 59.15 & 72.10 & 51.21 & 55.28 & 51.78 & 66.38 & 62.43 & 56.20 & 48.52 & 41.38 & 85.51 & 67.20 \\
    Fusing\cite{19ju2022fusing}{\tiny{} (ICIP2022)} & \textbf{100.00} & \underline{99.99} & 90.77 & \underline{99.99} & 74.14 & 55.93 & 76.50 & 78.63 & 76.49 & 58.50 & 78.67 & 62.49 & 72.51 & 67.33 & 88.23 & 78.68 \\    
    Univfd\cite{2ojha2023towards}{\tiny{} (CVPR2023)} & \textbf{100.00} & 99.97 & \textbf{99.88} & 99.80 & 99.23 & \textbf{99.98} & 53.64 & \underline{97.87} & 53.98 & 78.94 & 89.42 & 44.53 & 64.00 & 61.41 & 84.38 & 81.80 \\
    CLIPping\cite{CLIPping2024clipping}{\tiny{} (ICMR2024)} & 99.84 & 99.94 & 88.27 & 91.24 & 83.97 & 78.26 & 78.74 & 80.28 & 78.83 & 72.35 & 73.84 & 57.82 & 80.33 & 68.64 & 71.46 & 80.25 \\
    NPR\cite{npr2024rethinking}{\tiny{} (CVPR2024)} & \textbf{100.00} & \textbf{100.00} & 97.00 & \textbf{100.00} & 94.10 & 73.50 & 84.30 & 85.50 & 84.50 & 95.10 & \underline{92.00} & 78.90 & 83.40 & 60.10 & 86.40 & 87.65 \\
    Effort\cite{effort2024effort}{\tiny{} (ICML2025)} & \textbf{100.00} & \textbf{100.00} & 99.97 & 99.72 & 96.46 & 99.58 & \underline{94.95} & 98.73 & 97.54 & \textbf{96.62} & 90.25 & 78.62 & \textbf{98.26} & 83.63 & \textbf{88.87} & \underline{94.88} \\
    VIB-Net\cite{VIB2025towards}{\tiny{} (CVPR2025)} & \textbf{100.00} & \textbf{100.00} & 99.45 & 97.45 & 97.11 & 99.65 & 85.94 & 97.26 & 82.68 & 78.84 & 88.96 & 65.97 & 91.34 & \underline{85.93} & 83.77 & 90.29 \\
    \rowcolor{gray!20}
    \textbf{Ours} & \textbf{100.00} & \textbf{100.00} & \underline{99.52} & 99.88 & \underline{99.67} & \underline{99.94} & \textbf{96.92} & \textbf{98.83} & \textbf{96.72} & \underline{95.51} & \textbf{94.78} & 81.88 & \underline{97.86} & \textbf{92.88} & \underline{88.86} & \textbf{96.22} \\
    \bottomrule
    \end{tabular}%
  \label{tab1}%
\end{table*}%

\begin{table*}[t]
    \centering
    \renewcommand{\arraystretch}{1}
    \setlength{\tabcolsep}{2.5pt}
    \caption{Performance metrics (Accuracy) for various approaches trained using GAN-generated imagery. The statistics referenced in DIRE~\cite{26wang2023dire} are extracted from NPR~\cite{npr2024rethinking}, with "-" representing cases where the approach provided no data for the specific model under consideration.}    
    \begin{tabular}{ccccccccccccccccc}
    \toprule
    \multicolumn{1}{c}{\multirow{2}[4]{*}{Method}} & \multicolumn{6}{c}{Generative Adversarial Networks} & \multicolumn{7}{c}{Diffusion Models} & \multicolumn{2}{c}{Others} & \textit{ACC} \\
    \cmidrule(r){2-7} \cmidrule(r){8-14} \cmidrule(r){15-16} \cmidrule(r){17-17}    
    & \multicolumn{1}{p{2.5em}}{Pro-GAN} & \multicolumn{1}{p{2.5em}}{Style-GAN} & \multicolumn{1}{p{2.5em}}{Cycle-GAN} & \multicolumn{1}{p{2.5em}}{Star-GAN} & \multicolumn{1}{p{2.5em}}{Big-GAN} & \multicolumn{1}{p{2.5em}}{Gau-GAN} & SDV1.4 & VQDM & SDV1.5 & GLIDE & ADM & Midj & Wukong & \multicolumn{1}{p{2.5em}}{SAN} & \multicolumn{1}{p{2.5em}}{Deep-fake} & Avg \\
    \midrule
    FreDect\cite{23frank2020leveraging}{\tiny{} (ICML2020)} & \textbf{100.00} & \textbf{100.00} & 72.54 & 83.96 & 76.05 & 75.60 & 50.84 & 70.89 & 51.06 & 40.72 & 39.66 & 46.82 & 50.22 & 47.26 & 49.93 & 63.70 \\
    CNNSpot\cite{18wang2020cnn}{\tiny{} (CVPR2020)} & \textbf{100.00} & \textbf{100.00} & 79.25 & 91.99 & 71.57 & 80.56 & 53.15 & 60.81 & 53.83 & 53.80 & 62.19 & 48.03 & 51.41 & 55.25 & 54.56 & 67.76 \\
    Lgrad\cite{27tan2023learning}{\tiny{} (CVPR2023)} & 99.45 & 99.86 & 58.36 & 86.27 & 63.05 & 59.63 & 54.56 & 55.47 & 54.92 & 55.58 & 58.53 & 54.23 & 51.68 & 40.63 & 55.37 & 63.17 \\
    Fusing\cite{19ju2022fusing}{\tiny{} (ICIP2022)} & \underline{99.99} & \underline{99.98} & 79.15 & \underline{99.94} & 71.45 & 55.23 & 68.42 & 69.03 & 68.48 & 58.40 & 70.54 & 59.68 & 66.22 & 63.47 & 69.02 & 73.27 \\
    Univfd\cite{2ojha2023towards}{\tiny{} (CVPR2023)} & 99.59 & 99.94 & \underline{96.27} & 84.09 & \underline{93.32} & \underline{95.31} & 51.45 & \underline{91.34} & 51.52 & 69.22 & \underline{80.35} & 48.71 & 57.00 & 35.47 & 72.87 & 76.39 \\
    DIRE\cite{26wang2023dire}{\tiny{} (ICCV2023)} & \textbf{100.00} & 85.03 & 76.65 & \textbf{100.00} & 72.12 & 67.79 & 49.74 & 53.68 & 49.83 & 71.75 & 75.78 & 58.01 & 54.46 & - & - & - \\
    CLIPping\cite{CLIPping2024clipping}{\tiny{} (ICMR2024)} & 99.86 & 99.86 & 93.34 & 94.72 & 90.33 & 90.33 & 85.59 & 87.13 & 85.57 & 78.53 & 79.98 & 61.59 & 87.00 & \underline{72.83} & 72.83 & 85.30 \\
    NPR\cite{npr2024rethinking}{\tiny{} (CVPR2024)} & 99.70 & 99.80 & 88.60 & \textbf{100.00} & 83.00 & 73.00 & 74.00 & 77.00 & 74.00 & 78.00 & 81.00 & \textbf{71.00} & 72.00 & 63.00 & 53.00 & 79.14 \\
    Effort\cite{effort2024effort}{\tiny{} (ICML2025)} & \textbf{100.00} & \textbf{100.00} & 95.50 & 89.00 & 87.50 & 89.50 & \textbf{93.50} & 93.50 & \textbf{92.50} & \textbf{88.50} & 72.50 & 66.50 & \underline{91.00} & 74.50 & 73.00 & \underline{87.13} \\
    VIB-Net\cite{VIB2025towards}{\tiny{} (CVPR2025)} & 99.50 & 99.50 & 93.00 & 90.50 & 93.00 & \textbf{96.50} & 71.50 & 90.00 & 75.00 & 68.00 & 74.00 & 58.50 & 81.00 & 69.50 & \underline{73.50} & 82.20 \\
    \rowcolor{gray!20}
    \textbf{Ours} & \textbf{100.00} & \textbf{100.00} & \textbf{97.92} & 94.37 & \textbf{93.65} & 93.98 & \underline{91.39} & \textbf{94.79} & \underline{90.86} & \underline{87.72} & \textbf{86.39} & \underline{68.38} & \textbf{92.39} & \textbf{80.59} & \textbf{78.06} & \textbf{90.03} \\
    \bottomrule
    \end{tabular}%
  \label{tab2}%
\end{table*}%

\textit{2)GAN-trained Performance Evaluation:} With the continuous advancement of generative technology, an increasing number of unseen generative models are emerging, posing unprecedented challenges to real-versus-fake image detection. To verify the effectiveness of the proposed method when trained on GAN-based sources and evaluated on unseen models, we follow Protocol-2 to train and test all methods, with the results reported in Tab.~\ref{tab1} and~\ref{tab2}.

The experimental results reveal a clear performance divergence: almost all detection models perform excellently when dealing with variants within the GAN family. This can be attributed to the fact that GAN-generated images typically exhibit more pronounced artifact patterns than those produced by diffusion models, thus providing detectors with relatively easier and more explicit discriminative cues. However, when these models are applied to images generated by non-GAN models, their overall performance drops significantly. This phenomenon exposes a key issue: even if the training data contain rich features for distinguishing real from fake, the model may still fail to effectively generalize to unseen generative models. The underlying reason is that detectors often overfit to specific image content or pattern characteristics, which become invalid on unseen generated images and consequently degrade generalization. Although VIB-Net leverages the information bottleneck principle to mitigate overfitting to semantic content and shows certain advantages in unseen scenarios, its adaptation to specific generation patterns still unavoidably affects the accuracy of the decision boundary.

In contrast, the proposed method exhibits clear advantages in detecting various unseen generative models. Compared with Univfd, our method improves the average precision (AP) by 6.27\%. On unseen generative models (diffusion models), it achieves an 11.21\% accuracy gain over NPR. Compared with other competing methods, our approach yields a substantial improvement of 8\%–15\% in average precision. Furthermore, relative to the current state-of-the-art methods VIB-Net and Effort, our method improves accuracy by 2.9\%–7.8\%. These results validate the effectiveness of the proposed approach.

\begin{table*}[t]
    \centering
    \renewcommand{\arraystretch}{1}
    \setlength{\tabcolsep}{2.5pt}
    \caption{The AP results of various approaches trained on diffusion-based data are reported. Models not listed under the Diffusion Models (DMs) category are treated as out-of-distribution cases for evaluating cross-framework generalization.}    
    \begin{tabular}{ccccccccccccccccc}
    \toprule
    \multicolumn{1}{c}{\multirow{2}[4]{*}{Method}} & \multicolumn{6}{c}{Generative Adversarial Networks} & \multicolumn{7}{c}{Diffusion Models} & \multicolumn{2}{c}{Others} & \textit{AP} \\
    \cmidrule(r){2-7} \cmidrule(r){8-14} \cmidrule(r){15-16} \cmidrule(r){17-17}    
    & \multicolumn{1}{p{2.5em}}{Pro-GAN} & \multicolumn{1}{p{2.5em}}{Style-GAN} & \multicolumn{1}{p{2.5em}}{Cycle-GAN} & \multicolumn{1}{p{2.5em}}{Star-GAN} & \multicolumn{1}{p{2.5em}}{Big-GAN} & \multicolumn{1}{p{2.5em}}{Gau-GAN} & SDV1.4 & VQDM & SDV1.5 & GLIDE & ADM & Midj & Wukong & \multicolumn{1}{p{2.5em}}{SAN} & \multicolumn{1}{p{2.5em}}{Deep-fake} & Avg \\
    \midrule
    Univfd{\tiny{} (CVPR2023)} 
    & 96.17 & 77.48 & 97.69 & 86.84 & 85.42 & 92.38 & 93.36 & 98.04 & 96.81 & \underline{96.78} & 92.74 & \underline{94.30} & 93.71 & 92.41 & 79.22 & 91.56 \\
    NPR{\tiny{} (CVPR2024)}
    & 88.27 & 87.46 & 81.10 & \textbf{100.00} & 81.27 & 86.07 & 79.99 & 85.65 & 79.18 & 71.56 & 61.58 & 60.8 & 83.19 & 72.30 & \textbf{92.82} & 80.75 \\
    CLIPping{\tiny{} (ICMR2024)}
    & 92.86 & 92.24 & 96.66 & 80.00 & 87.53 & 93.02 & 98.03 & 95.74 & 97.69 & \textbf{98.31} & 80.09 & 83.57 & 95.82 & 93.97 & 54.31 & 89.32 \\
    Effort{\tiny{} (ICML2025)}
    & \underline{97.97} & 92.11 & \underline{97.98} & \underline{97.08} & 89.17 & 92.61 & \underline{99.88} & \underline{99.97} & 97.99 & 94.70 & 92.30 & 90.24 & \underline{97.98} & 92.72 & 76.21 & 93.93 \\
    VIB-Net{\tiny{} (CVPR2025)}
    & \textbf{98.66} & \underline{94.85} & \textbf{98.95} & 96.67 & \underline{91.08} & \underline{94.12} & \textbf{100.00} & \textbf{100.00} & \underline{98.46} & 95.95 & \underline{94.00} & 92.47 & 98.91 & \underline{97.72} & 71.12 & \underline{94.86} \\
    \rowcolor{gray!20}
    \textbf{Ours} 
    & 96.89 & \textbf{96.70} & 95.87 & 96.22 & \textbf{94.84} & \textbf{96.81} & \textbf{100.00} & \textbf{100.00} & \textbf{99.96} & 95.89 & \textbf{94.78} & \textbf{95.91} & \textbf{99.93} & \textbf{98.19} & \underline{82.85} & \textbf{96.32} \\
    \bottomrule
    \end{tabular}%
  \label{tab3}%
\end{table*}%

\begin{table*}[t]
    \centering
    \renewcommand{\arraystretch}{1}
    \setlength{\tabcolsep}{2.5pt}
    \caption{The comparison is conducted using classification accuracy, which depends on the chosen decision threshold. To ensure fairness, all methods adopt a unified threshold of 0.5.}    
    \begin{tabular}{ccccccccccccccccc}
    \toprule
    \multicolumn{1}{c}{\multirow{2}[4]{*}{Method}} & \multicolumn{6}{c}{Generative Adversarial Networks} & \multicolumn{7}{c}{Diffusion Models} & \multicolumn{2}{c}{Others} & \textit{ACC} \\
    \cmidrule(r){2-7} \cmidrule(r){8-14} \cmidrule(r){15-16} \cmidrule(r){17-17}    
    & \multicolumn{1}{p{2.5em}}{Pro-GAN} & \multicolumn{1}{p{2.5em}}{Style-GAN} & \multicolumn{1}{p{2.5em}}{Cycle-GAN} & \multicolumn{1}{p{2.5em}}{Star-GAN} & \multicolumn{1}{p{2.5em}}{Big-GAN} & \multicolumn{1}{p{2.5em}}{Gau-GAN} & SDV1.4 & VQDM & SDV1.5 & GLIDE & ADM & Midj & Wukong & \multicolumn{1}{p{2.5em}}{SAN} & \multicolumn{1}{p{2.5em}}{Deep-fake} & Avg \\
    \midrule
    Univfd{\tiny{} (CVPR2023)} 
    & 83.00 & 64.50 & 85.50 & 57.50 & 71.50 & 77.50 & 91.50 & 93.00 & 90.00 & 92.00 & 86.00 & \textbf{87.50} & 88.50 & 82.50 & 51.00 & 80.10 \\
    NPR{\tiny{} (CVPR2024)} 
    & 74.25 & 69.17 & 73.77 & \textbf{98.87} & 63.85 & 65.98 & 85.58 & 86.87 & 84.75 & 75.53 & 63.31 & 63.29 & 85.03 & 70.09 & 61.05 & 74.76 \\
    CLIPping{\tiny{} (ICMR2024)} 
    & 92.10 & \underline{92.01} & 93.22 & 87.42 & \textbf{85.48} & 86.77 & 97.55 & 95.23 & 96.32 & \textbf{96.92} & 79.80 & \underline{82.08} & 93.47 & \underline{92.66} & 57.98 & \underline{88.60} \\
    Effort{\tiny{} (ICML2025)} 
    & \textbf{95.50} & 89.00 & \underline{96.50} & 82.50 & 80.50 & 86.00 & 97.00 & 96.50 & \underline{97.00} & 90.00 & 88.10 & 76.00 & \underline{97.50} & 88.50 & \underline{61.50} & 88.14 \\
    VIB-Net{\tiny{} (CVPR2025)} 
    & \underline{95.00} & 91.00 & \textbf{97.00} & 71.00 & 73.00 & \underline{87.50} & \underline{99.00} & \underline{98.50} & \underline{97.00} & \underline{93.50} & \textbf{92.50} & 80.00 & \underline{97.50} & \textbf{94.00} & 49.50 & 87.73 \\
    \rowcolor{gray!20}
    \textbf{Ours} 
    & 93.50 & \textbf{92.66} & 94.03 & \underline{91.00} & \underline{84.08} & \textbf{89.45} & \textbf{99.96} & \textbf{99.99} & \textbf{99.28} & 93.22 & \underline{92.29} & 72.54 & \textbf{98.62} & 89.27 & \textbf{72.64} & \textbf{90.84} \\
    \bottomrule
    \end{tabular}%
  \label{tab4}%
\end{table*}%

\textit{3)Diffusion-trained Performance Evaluation:} The diffusion model family can generate images with an extremely high level of visual realism, which poses substantial challenges for real-vs-fake image detection. To verify the effectiveness of our method when dealing with high-fidelity generated content, we consider a challenging experimental setting: the model is trained on images generated by Stable Diffusion v1.4 and VQDM, and then evaluated on 15 different generative models (including the latest diffusion models and a series of GAN-based models). The detailed experimental results are shown in Tab.~\ref{tab3} and~\ref{tab4}.

Most detection models exhibit a pronounced performance degradation on high-quality images generated by diffusion models. In particular, for generators such as ADM, Midjourney, and VQDM, the accuracies of detectors like Univfd and Lgrad drop by around 40\%. We attribute this performance degradation to two key factors: on the one hand, the high-quality images produced by advanced generative models make it difficult for traditional methods to extract effective discriminative features; on the other hand, existing detectors tend to overfit pattern- or content-related biases tied to the training distribution rather than capturing intrinsic forensic cues. As a result, they perform relatively well on models with distributions similar to the training data (e.g., SDv1.5, Wukong), but exhibit notably insufficient generalization to models whose generated pattern or content differs more substantially. It is worth noting that VIB-Net leverages the information bottleneck principle to decouple and filter out irrelevant semantic features, thereby reducing dependence on specific content to some extent. Effort expands the CLIP feature space via orthogonal bases to prevent overfitting to particular patterns and also achieves competitive performance. However, both methods still show certain limitations when applied to Deepfake-style datasets with highly homogeneous content. In the experiments on GAN-based generators, all competing methods also perform poorly overall—especially on models such as BigGAN and GauGAN, where the accuracy typically hovers around 50\%–60\%, further confirming the constraints imposed by asymmetric bias learning on model generalization.

In contrast, our method achieves the best performance across all evaluation metrics. This demonstrates that, by introducing an adversarial mechanism, our approach effectively avoids overfitting to generative-pattern and content biases and encourages the model to focus on more essential and shared characteristics of generated images, thereby substantially enhancing its generalization capability. Overall, the experimental results clearly validate the effectiveness and superiority of the proposed method.

\begin{table}[t]
\caption{Ablation Study Results}
\label{tab:ablation}
\centering
\renewcommand{\arraystretch}{1}
\begin{tabular}{ccccccc}
\hline
CLIP:ViT & AFL & EMLoss & FALoss & LRLoss & Acc & AP \\
\hline
\checkmark & & & & & 82.17 & 88.40 \\
\checkmark & \checkmark & & & & 85.30 & 92.31 \\
\checkmark & \checkmark & \checkmark & \checkmark & & 87.67 & 94.70 \\
\checkmark & \checkmark & \checkmark & & \checkmark & 85.40 & 93.48 \\
\checkmark & \checkmark & & \checkmark & \checkmark & 87.71 & 94.59 \\
\checkmark & \checkmark & \checkmark & \checkmark & \checkmark & \textbf{90.03} & \textbf{96.22} \\
\hline
\end{tabular}
\label{tab_ablation}%
\end{table}

\subsection{Ablation Study}

To further validate the effectiveness of the adversarial mechanism in MAFL and the independent contribution of each loss component, we conduct a systematic ablation study under Protocol-2. The compared variants are summarized as follows:

\begin{itemize}
\item[$\bullet$] Baseline: The basic detector trained only with CLIP (ViT-L/14) features and a binary classifier.
\item[$\bullet$] DA-AFL: Extends Baseline with conventional symmetric domain-adversarial training (e.g., DANN) to examine the impact of generic adversarial feature learning (AFL) in AIGI detection.
\item[$\bullet$] EF-AFL: Combines Entropy Maximization Loss (EMLoss) and Feature Alignment Loss (FALoss), enforcing asymmetric constraints from distribution and feature-structure perspectives.
\item[$\bullet$] EL-AFL: Combines EMLoss and Label Reversal Loss (LRLoss), enforcing asymmetric constraints from distribution and decision boundary perspectives.
\item[$\bullet$] Uses FALoss and LRLoss to constrain feature learning jointly from structure and decision perspectives.
\item[$\bullet$] The complete framework with all three losses, jointly suppressing distribution–structure–decision biases.
\end{itemize}

\textit{1) DA-AFL:} As shown in Tab.~\ref{tab_ablation}, DA-AFL improves accuracy from 82.17\% to 85.30\% and AP from 88.40\% to 92.31\%. This demonstrates that adversarial feature learning can alleviate overfitting to generator-specific patterns. However, symmetric domain alignment offers only single-dimension, weak constraints and thus remains limited for this task.

\textit{2) EF-AFL:} Replacing symmetric domain adversarial loss with EMLoss and FALoss further boosts accuracy to 87.67\% and AP to 94.70\%. This confirms the necessity of jointly suppressing generator-dependent statistical deviations and enforcing manifold consistency across fake images—capabilities absent in DANN-style training.

\textit{3) EL-AFL:} Removing FALoss leads to reduced performance (85.40\%/93.48\%), indicating that without structural consistency, the latent space becomes fragmented by generator-specific characteristics, undermining the generalization benefit of decision-level adversarial learning.

\textit{4) FL-AFL:} Applying structure + decision constraints maintains strong performance (87.71\%/94.59\%), highlighting the effectiveness of structural alignment and the stronger adversarial pressure produced by decision-level perturbation. Yet, lacking distribution equalization still leaves residual pattern biases.

\textit{5) MAFL(Full):} The full model achieves the best performance, with 90.03\% accuracy and 96.22\% AP. This demonstrates that the three losses are highly complementary—only multi-dimensional, asymmetric adversarial suppression can fully mitigate generator-specific biases and maximize cross-model generalization.

\subsection{Performance with Limited Training Data}

\begin{figure}[t]
	\centering
		\includegraphics[width=1\linewidth]{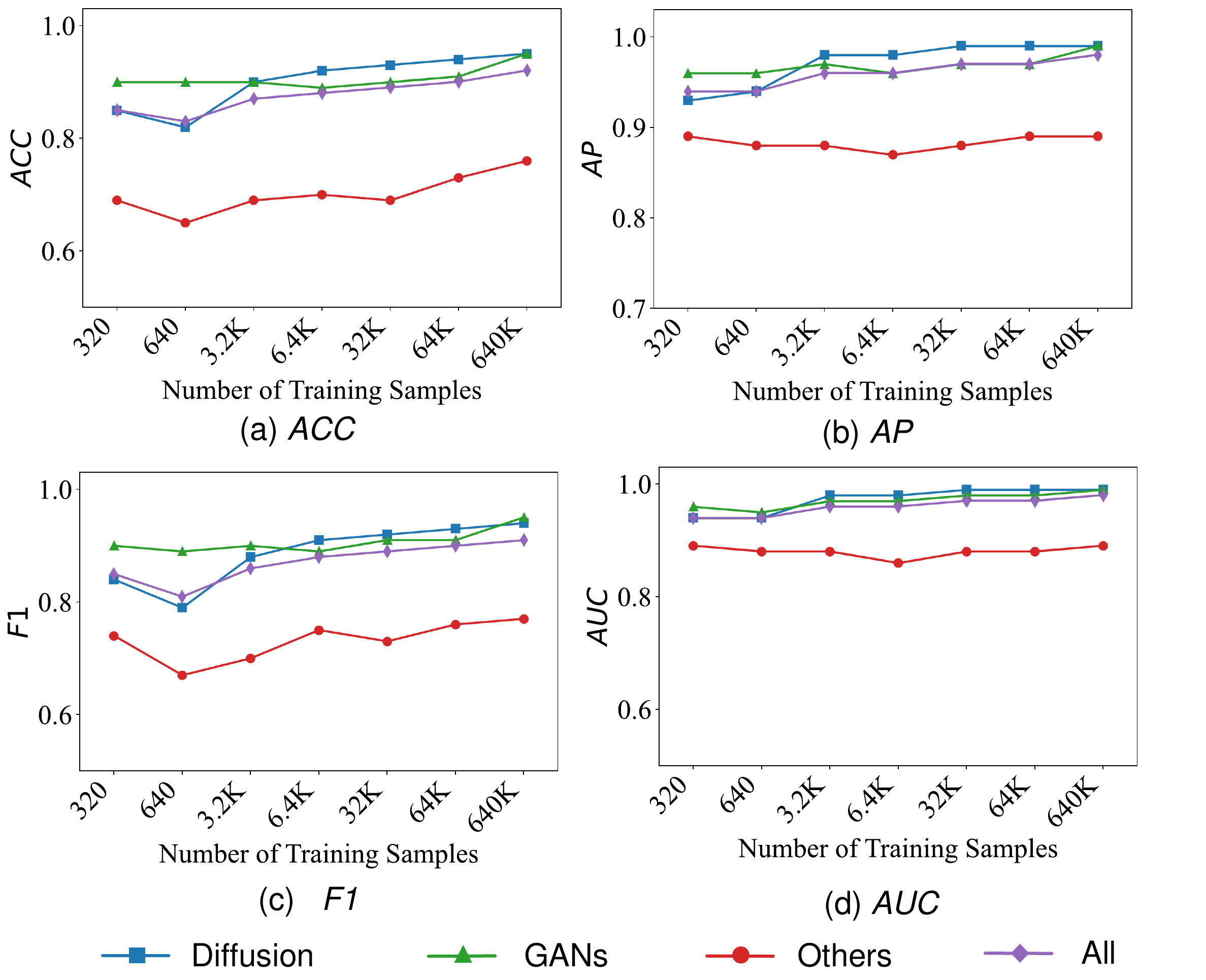}
	\caption{Test results with different scales of training samples. Our method can still achieve high detection performance when trained with only 300-700 images.}
	\label{FIG:6}
\end{figure}

In real-world scenarios, authentic and generated images are often difficult to obtain in large quantities as in controlled experiments, making performance evaluation under limited data especially important. To this end, following Protocol-3, we randomly downsampled subsets of different sizes (64K, 32K, 6.4K, 3.2K, 640, and 320 images) from the full 640K training set and evaluated performance across various types of generative models, as shown in Fig.~\ref{FIG:6}. Notably, even with extremely small amounts of training data, our method still maintains strong detection capability—using only 320 training images, the model achieves over 80\% ACC and 90\% AP on both diffusion models and GANs. Performance is slightly lower on other forgery types (e.g., SAN and Deepfake), likely because these methods focus mainly on facial manipulation with relatively limited variation in pattern and content, thereby reducing the comparative advantage of our method. Overall, MAFL demonstrates excellent robustness under data-scarce conditions: even when the training data is reduced by three orders of magnitude, it still maintains strong detection and generalization performance.

\subsection{Robustness Evaluation Against Perturbations}

\begin{figure}[t]
	\centering
		\includegraphics[width=0.95\linewidth]{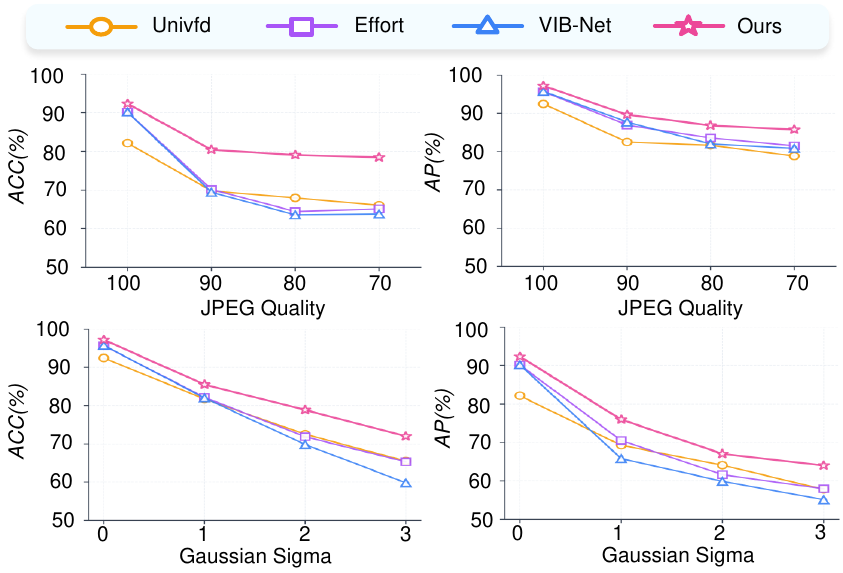}
	\caption{Robustness evaluation of different methods under perturbations in Protocol-3, with the top row showing the effects of JPEG compression and the bottom row showing the effects of Gaussian blur.}
	\label{FIG:7}
\end{figure}

In real-world scenarios, images often undergo various post-processing operations. Therefore, in addition to focusing on the detector’s cross-model generalization ability and performance under few-shot settings, it is also crucial to evaluate its robustness under different perturbations. In this experiment, we follow the Protocol-3 setup to assess the robustness of multiple methods. During testing, samples are subjected to two common post-processing operations: JPEG compression and Gaussian blur. The perturbation strength is controlled by JPEG quality factors {90, 80, 70} and Gaussian blur parameters {1, 2, 3}. The experimental results are shown in Fig.~\ref{FIG:7}, where ACC and AP are reported for all methods.

We observe that as the perturbation strength increases, most existing methods exhibit a significant performance drop. In contrast, our method consistently achieves the highest ACC and AP across all levels of perturbation, demonstrating stronger robustness to compression and blurring distortions. This robustness benefits from the introduced adversarial learning strategy, which encourages the model to focus on multi-dimensional, generator-shared generative features across different generators and to establish an adversarial game between spurious correlations and truly discriminative features, rather than relying on single-dimensional or pixel-level anomalies.

\subsection{Discussion on Training Sources}

\begin{figure}[t]
	\centering
		\includegraphics[width=1\linewidth]{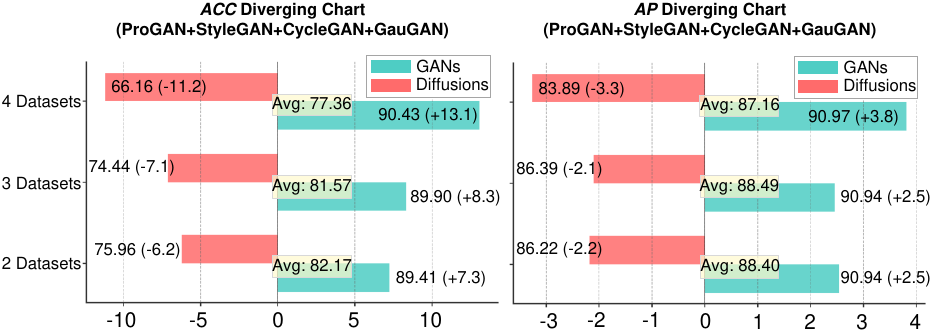}
	\caption{Training the baseline model (CLIP + classifier) using 2 to 4 datasets. 2: ProGAN + StyleGAN; 3: ProGAN + StyleGAN + CycleGAN; 4: ProGAN + StyleGAN + CycleGAN + GauGAN}
	\label{FIG:8}
\end{figure}

\begin{figure}[t]
	\centering
		\includegraphics[width=1\linewidth]{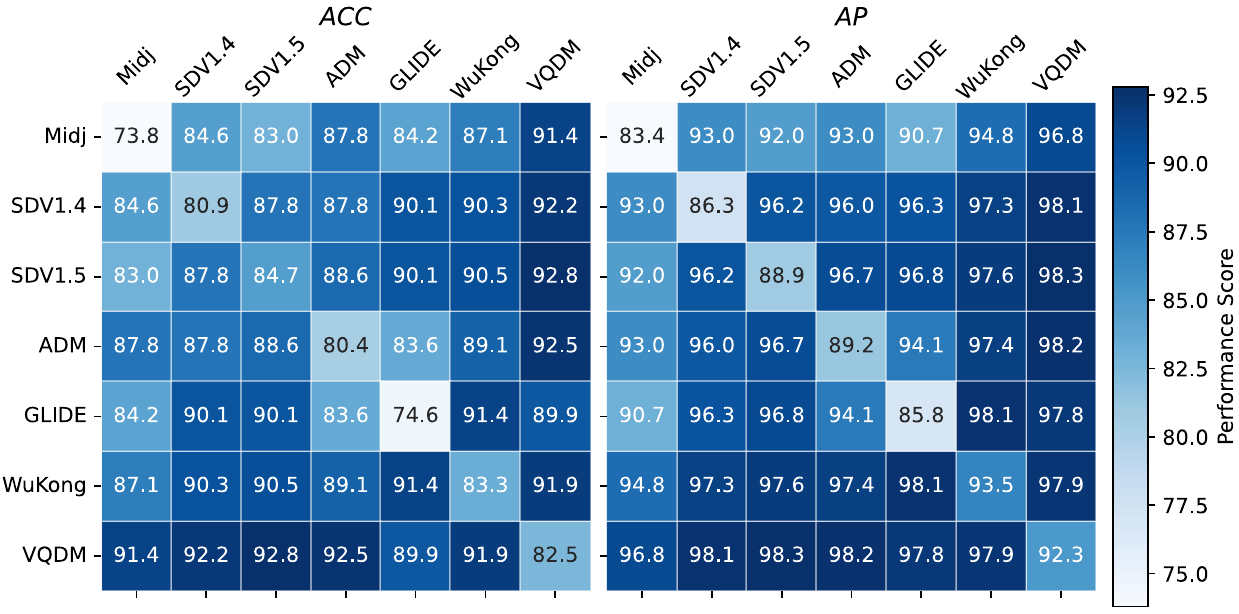}
	\caption{Test results of training MAFL with pairwise combinations of 7 generative models. Training with the same generative model (diagonal) shows poor performance.}
	\label{FIG:9}
\end{figure}

Due to the data-driven nature of deep learning, the training set has a significant impact on the model. Therefore, in this section, we discuss how the choice of the types and number of generative models during training affects detector performance. First, to evaluate the influence of the number of generative models in the training set and to assess the effectiveness of our method compared with the baseline (CLIP:ViT-L/14 + classifier), we train the baseline model using 2, 3, and 4 generative models respectively, and test it on other generative models. The experimental results are shown in Fig.~\ref{FIG:8}. We observe an interesting phenomenon: as the number of GAN types increases, the model’s performance on GANs improves, but it drops significantly on diffusion models. This is likely due to overfitting to the pattern and content characteristics of GANs, which further validates the effectiveness of our method. Therefore, we can conclude that, for the model, seeing more images generated by similar-distribution generators during training may cause it to over-learn the generation style or artifact patterns of that family, which is detrimental to generalization to other generator families.

Furthermore, we investigate the impact of the choice of generative model types on detector performance. Specifically, we use different combinations of generative models as the training set to train MAFL, and the experimental results are shown in Fig.~\ref{FIG:9}. It can be observed that when two generative models are used for training, the method achieves superior performance, especially with the combination of SDV1.4 and VQDM. Avoiding overfitting to the patterns of these two generative models helps improve detection of unseen generative models. In contrast, training with only a single generative model (diagonal entries) leads to noticeably lower performance.

\subsection{Effect of MAFL with different backbone}

\begin{figure*}[htbp]
	\centering
		\includegraphics[width=1\linewidth]{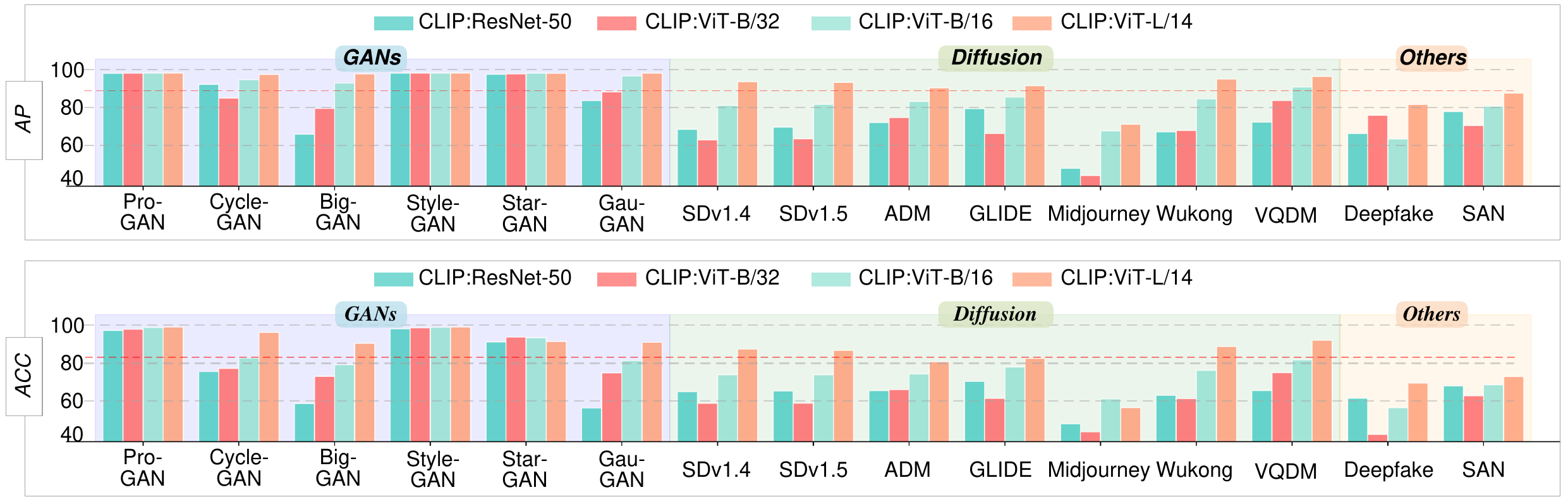}
	\caption{Performance comparison of different MAFL variants. Using CLIP:ViT-L/14 as the backbone achieved the highest performance.}
	\label{FIG:10}
\end{figure*}

So far, we have observed the impressive performance achieved by the adversarial feature learning framework (MAFL) when combined with the ViT-L/14 backbone. In this section, we investigate the effect of transferring MAFL to different backbone networks. We consider the following variants to construct MAFL: (1) CLIP:ResNet50, (2) CLIP:ViT-B/16, (3) CLIP:ViT-B/32, and (4) CLIP:ViT-L/14. For each MAFL variant, we conduct experiments following Protocol-2, and the results are shown in Fig.~\ref{FIG:10}. It can be observed that MAFL also performs well when integrated with other backbone variants, demonstrating strong adaptability. Among them, the combination with ViT-L/14 achieves the best performance, which can be attributed to its larger model capacity and its attention-based global image understanding capability.

\subsection{Analysis of Features Extracted by MAFL}

Finally, we apply the proposed method to address the issues observed in Sec.~\ref{Observations and Motivations}. In Sec.~\ref{Observations and Motivations}, we pointed out that CLIP features are highly sensitive to semantic content (Fig.~\ref{FIG:2}(a)) and, during real/fake classification training, tend to capture model-specific generative patterns (Fig.~\ref{FIG:2}(c)), which degrades generalization performance. As a solution, we introduced MAFL.

Under the same experimental setting as in Sec.~\ref{Observations and Motivations}, we visualize the features extracted by MAFL and obtain encouraging results, as shown in Fig.~\ref{FIG:2}(b)(d). We observe that the MAFL-optimized CLIP features are no longer sensitive to content bias or generative-pattern bias. All clusters are correctly separated according to real vs. generated labels, rather than by content or generative model. This validates the effectiveness of our method.

\subsection{Discussion on the limitations of the developed methods}

A long-standing core issue in deep-learning-based AI-generated image detection is that, in real-world applications, the detector is expected not only to output a binary authenticity decision, but also to provide human-understandable and trustworthy explanations. However, due to the highly black-box nature of neural network decision processes, the lack of interpretability makes it difficult for detection results to serve as reliable evidence in scenarios such as judicial forensics and media authentication. To address this, a promising direction is to introduce the Chain-of-Thought (CoT) reasoning capability of large models, enabling multimodal models to progressively analyze generative traces along an explicit reasoning path during detection, thereby improving detection performance while simultaneously offering clear and intuitive explanatory evidence.

Another important challenge is that existing methods often remain confined to authenticity discrimination and largely overlook the provenance and attribution of generated images. From the perspective of supervised learning, authenticity detection mainly relies on the common features shared by generated images, whereas source attribution depends on the specific characteristics associated with different generative models and their mechanisms. Therefore, it is necessary to explore more extensible task designs, such as constructing hierarchical multi-task learning frameworks that jointly model both common and specific feature spaces, thus enabling a transition from authenticity verification to fine-grained attribution at the level of generative models and even specific model versions.

\section{Conclusion}
\label{sec:Conclusion}

We propose a Multi-dimensional Adversarial Feature Learning (MAFL) method to address the asymmetric bias learning problem in generative image detection, where models tend to overfit specific generative patterns or content. This bias causes misclassification on unseen generative patterns or content, limiting generalization. To overcome this, we adopt an adversarial feature learning strategy that prevents overfitting to specific pattern and content, guiding the model to focus on essential common features of generated images and thereby improving generalization. In the experiments, we designed multiple evaluation protocols based on a wide range of generative model types and families, including GANs, diffusion models, VAEs, and flow-based models. Compared with recent state-of-the-art detection methods such as VIB-Net and Effort, our approach demonstrates notable advantages in terms of generalization, limited-data scenarios, and robustness, thus validating the effectiveness of the proposed framework. Furthermore, ablation studies confirm the independent contributions of each component. We believe that this framework, with its strong extensibility and performance, can serve as a promising solution for scalable AI-generated image detection.

\bibliographystyle{IEEEtran}
\bibliography{references}

\end{document}